\documentclass[11pt]{article}

\PassOptionsToPackage{hyperfootnotes=false}{hyperref}
\usepackage{acl}

\usepackage{times}
\usepackage{latexsym}

\usepackage[T1]{fontenc}

\usepackage[utf8]{inputenc}

\usepackage{microtype}

\usepackage{inconsolata}

\usepackage{graphicx}
\usepackage{booktabs}
\usepackage{multirow}
\usepackage{amsmath}
\usepackage{amssymb}
\usepackage{xcolor}
\usepackage{tcolorbox}
\tcbuselibrary{breakable}
\usepackage[ruled,vlined]{algorithm2e}
\title{Decoupling Search from Reasoning: A Vendor-Agnostic Grounding Architecture for LLM Agents}

\author{\textbf{Emmanuel Aboah Boateng, Kyle MacDonald, Amardeep Kumar,} \\
  \textbf{Siddharth Kodwani, Sudeep Das} \\
  DoorDash, Inc. \\
  \texttt{\{e.aboahboateng, kyle.macdonald, amardeep.kumar,} \\
  \texttt{siddharth.kodwani, sudeep.das2\}@doordash.com}}

\begin{document}
\maketitle
\begin{abstract}
Production LLM agents increasingly depend on real-time search, yet native search grounding bundles retrieval policy, provider choice, evidence injection, cost, latency, and generation behavior behind a single model-provider boundary. This coupling makes grounding hard to inspect, tune, reuse, or port, and can trigger Search-Induced Verbosity that breaks strict output contracts. We present Decoupled Search Grounding (DSG), a vendor-agnostic boundary that moves grounding outside the reasoning model through an MCP-compatible gateway, exposing provider routing, source-aware context rendering, configured fallback, retrieval-depth control, and exact plus semantic caching as first-class controls. Across five frontier models on SimpleQA, FreshQA, and HotpotQA, native search leads on recency-sensitive FreshQA, but DSG exposes a stronger frontier when control matters: on SimpleQA it nearly matches native accuracy (86.1\% vs.\ 87.7\%) at 91\% lower search cost, preserves concise answer contracts, and reaches a 99.4\% warm-cache hit rate with 68\% lower latency. Deployed as a shared production grounding layer for large-scale agentic workloads with interchangeable models, DSG matches native-search accuracy on an e-commerce query-understanding (QIU) workload while cutting search cost by over 98\%. Real-time grounding is best treated as an optimizable interface boundary, not a fixed model feature.
\end{abstract}

\section{Introduction}
\label{sec:introduction}

\begin{figure}[!t]
    \centering
    \includegraphics[width=\linewidth]{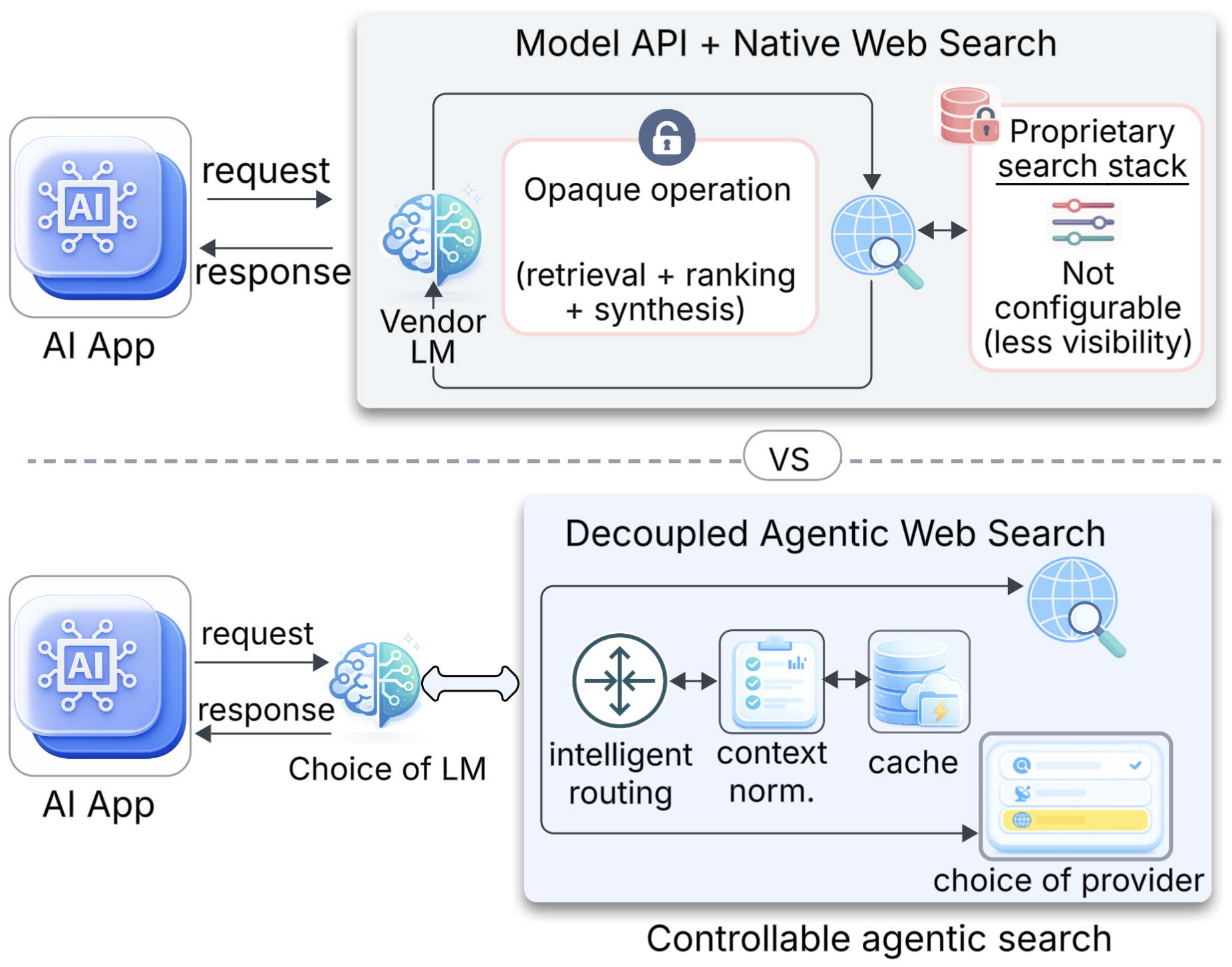}
    \caption{Decoupled Search Grounding (DSG) separates retrieval from model-native generation, making provider choice, caching, and output control explicit.}
    \label{fig:interface_shift}
\end{figure}

The deployment of Large Language Models (LLMs) in production increasingly centers on agentic tasks: multi-step workflows in which a model reasons over intermediate state, decides when to call external tools, and often uses search to ground actions in fresh evidence \citep{schick2024toolformer, mialon2023augmented, yao2022react, li2023apibank, qin2024toolllm}. These systems must satisfy strict cost, latency, reliability, and output-contract constraints. As static Retrieval-Augmented Generation (RAG) pipelines evolve into dynamic, tool-using workflows \citep{lewis2020retrieval, karpukhin2020dpr, izacard2021fid, jiang2023flare, asai2024selfrag}, the interface between the reasoning model and real-time search becomes a first-order systems decision. Deployed NLP studies document practical trade-offs in retrieval quality, latency, dynamic knowledge maintenance, and model efficiency \citep{murtaza2025implementing, chen2025comrag, jucla2026retrieval, li2025proactive}.

A common integration path is \textit{native search grounding}: a model-provider feature in which the model API invokes a provider-managed search stack and injects retrieved evidence into generation. Native integrations are convenient and often strong, especially on recency-sensitive questions. Teams can instead integrate external providers directly, but their APIs, pricing, result formats, ranking, and failure surfaces differ substantially, which pushes engineers toward the default native path even when production systems need explicit control over provider choice and retrieval behavior. This coupling hides decisions that production teams commonly need to control: which provider is queried, how much context is retrieved, how costs are bounded, whether repeated queries can be cached, and how tool outputs interact with downstream schema constraints.

We study this interface boundary directly. Prior work documents uncertainty-associated verbosity and studies over-searching in search-augmented systems \citep{zhang2025verbosity_compensation, xie2026oversearching}; here we isolate an interface-specific failure mode, \textit{Search-Induced Verbosity}, in which native search changes response style despite strict output instructions. Strict prompts such as ``Provide only the final answer entity'' can yield explanatory paragraphs, which matters for production systems where LLMs frequently act as intermediate nodes whose outputs are parsed, classified, cached, or routed by downstream services.

We propose a vendor-agnostic Decoupled Search Grounding (DSG) architecture implemented through an MCP-compatible gateway. DSG gives applications and reasoning models a stable interface while provider-specific configuration and adapters remain behind the gateway. Instead of treating search as an opaque model feature, DSG exposes search as a structured tool layer that can normalize provider outputs, route requests, cache repeated queries, and preserve the separation between retrieval and reasoning. The key contribution is the controllable boundary itself: DSG externalizes provider policy, cache reuse, fallback, and source-aware context rendering while keeping the reasoning model interchangeable.

We evaluate DSG against native search across five frontier models (GPT-4o, GPT-4o-mini, Gemini 2.5 Flash, Gemini 2.5 Pro, and Claude Sonnet 4), public QA benchmarks (SimpleQA, FreshQA, HotpotQA), and proprietary e-commerce Query Intent Understanding (QIU) datasets. Our contributions are:

\begin{enumerate}
    \item \textbf{A systems boundary for grounding:} We formalize search grounding as a decoupled, provider-agnostic layer that keeps the reasoning model interchangeable while exposing retrieval policy, caching, and provider choice as explicit system controls.
    \item \textbf{A search-intelligence and standardization framework:} We implement DSG as an MCP-compatible gateway that gives agents a stable search interface, normalizes heterogeneous provider outputs into source-aware context, and supports cache-gated retrieval with configured provider fallback across search backends.
    \item \textbf{Evidence on behavior, quality, and operations:} We identify Search-Induced Verbosity as a prompt-compliance risk in native search grounding and show how DSG's structured tool boundary helps mitigate it, while quantifying accuracy, cost, latency, retrieval-depth, and cache-reuse trade-offs across public QA benchmarks and a production workload.
\end{enumerate}

DSG makes grounding an interface decision: provider choice, cost policy, cache reuse, context rendering, fallback, and output-contract behavior become first-class controls while the reasoning model stays interchangeable. Native and decoupled search expose different trade-offs; the value of DSG is exposing that choice as an explicit, controllable decision.

\section{Related Work}
\label{sec:related_work}

Toolformer \citep{schick2024toolformer}, ReAct \citep{yao2022react}, API-Bank \citep{li2023apibank}, ToolLLM \citep{qin2024toolllm}, and WebGPT \citep{nakano2021webgpt} establish that LLMs can use external tools and APIs. These works primarily study agent capability and benchmark performance; we focus instead on the production consequences of where the search interface sits in the model stack.

RAG methods improve knowledge-intensive question answering and, in some settings, factual generation \citep{lewis2020retrieval, karpukhin2020dpr, izacard2021fid}, with later work exploring dynamic retrieval, retrieval-reasoning interaction, and retrieval-backed revision, including FLARE \citep{jiang2023flare}, RARR \citep{gao2023rarr}, Self-RAG \citep{asai2024selfrag}, Adaptive-RAG \citep{jeong2024adaptive}, and IRCoT \citep{trivedi2023ircot}. Separate research threads study prompt optimization \citep{zhou2023ape} and strong-to-weak model adaptation \citep{hsieh2023distilling, boateng2025strongweak, boateng2025conceptdistill}. DSG is complementary to these methods because it stabilizes the grounding interface as prompts and reasoning models change. Industry applications such as multi-category e-commerce intent understanding further motivate multi-source grounding \citep{boateng2026agentic}. Our contribution is orthogonal: we study whether search is embedded as opaque provider behavior or exposed as an explicit, controllable subsystem.

Our evaluation combines short-form factuality (SimpleQA; \citealp{wei2024simpleqa}), multi-hop stress testing (HotpotQA; \citealp{yang2018hotpotqa}), and recency-sensitive grounding (FreshQA/FreshLLMs; \citealp{vu2024freshllms}). For scalable assessment, we draw on LLM-as-judge and RAG evaluation work including G-Eval \citep{liu2023geval}, MT-Bench / Chatbot Arena analysis \citep{zheng2023judging}, ARES \citep{saadfalcon2024ares}, and RAGAs \citep{es2024ragas}.

Industry-track work increasingly documents deployed retrieval trade-offs in enterprise support, dynamic vector stores, and latency-constrained search \citep{murtaza2025implementing, jucla2026retrieval, chen2025comrag, li2025proactive}. Early MCP research examines tool-description quality and proposes production deployment patterns \citep{mcp_smelly_2026, mcp_patterns_2026}. We use MCP as the connection substrate for DSG, and evaluate decoupled search grounding against proprietary native search. Distinct from these lines of work, we treat real-time search grounding itself as a controllable, vendor-agnostic interface and measure the cost, latency, caching, and output-contract trade-offs it exposes against proprietary native search across multiple frontier models and a production workload.
\section{The Challenge of Native Search}
\label{sec:challenge}

Native search is provider-managed grounding: the model API invokes search and injects evidence through provider-specific context handling. This one-call abstraction is attractive for agents, but it moves retrieval policy inside the provider boundary. Developers lose control over provider selection, result normalization, caching, tool-output formatting, and migration paths when a better reasoning model or search provider becomes available.

When the model call is an intermediate step in an automated pipeline, this coupling creates four deployment challenges: model-provider lock-in, opaque latency, fixed cost structures, and instruction-following degradation that we term \textit{Search-Induced Verbosity}.

\begin{figure}[t]
    \centering
    \includegraphics[width=\linewidth]{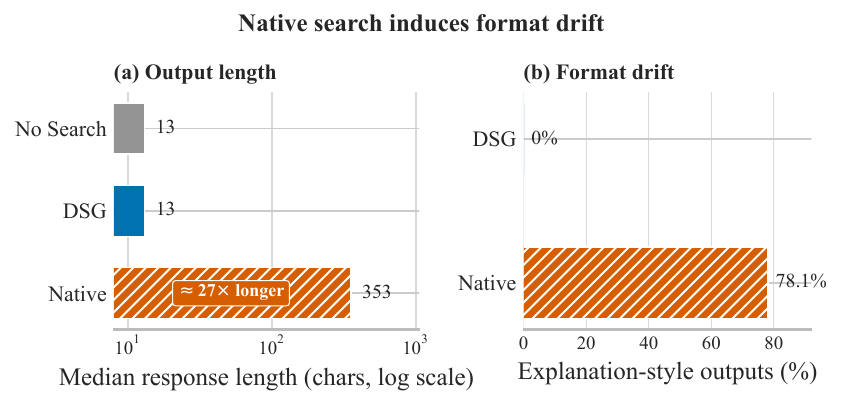}
    \caption{Prompt-compliance diagnostic on HotpotQA. Native search can shift concise answer extraction into explanatory output, while DSG preserves the answer-only interface required by downstream parsers.}
    \label{fig:verbosity}
\end{figure}

\subsection{Search-Induced Verbosity}
Production pipelines often use LLMs as intermediate reasoning nodes whose outputs must parse as JSON, boolean flags, or short entities. Prompts therefore specify strict output contracts, such as ``Provide ONLY the final answer entity.''

We observe that native search can shift the model from concise extraction to explanatory summarization, even when the final-answer instruction is explicit. The answer may be present, but extra prose can break exact-match evaluation or downstream parsers.

\begin{tcolorbox}[
    colback=white,
    colframe=gray!55!black,
    colbacktitle=gray!12!white,
    coltitle=black,
    title=\textbf{Example: Prompt-Compliance Failure},
    boxrule=0.5pt,
    arc=1mm,
    left=4pt,
    right=4pt,
    top=4pt,
    bottom=4pt
]
\footnotesize
\textbf{Instruction:} \textit{``Provide ONLY the final answer entity.''}\\
\textbf{Question:} \textit{``Is the building at 200 West Street taller than 888 7th Avenue?''}

\vspace{1mm}
\begin{tcolorbox}[colback=red!4!white,colframe=red!60!black,boxrule=0.4pt,arc=1mm,left=3pt,right=3pt,top=2pt,bottom=2pt]
\textbf{\color{red!70!black}Native search output:}
\textit{``Based on the search results, I can compare the heights of both buildings: 200 West Street is 749 feet tall... while 888 7th Avenue is 628 feet tall...''}
\hfill \textbf{245 chars}
\end{tcolorbox}

\vspace{0.5mm}
\begin{tcolorbox}[colback=green!4!white,colframe=green!45!black,boxrule=0.4pt,arc=1mm,left=3pt,right=3pt,top=2pt,bottom=2pt]
\textbf{\color{green!45!black}DSG output:}
\textit{``Yes''}
\hfill \textbf{3 chars}
\end{tcolorbox}
\end{tcolorbox}

We treat this as a deployment-risk diagnostic, not as a claim that native search always changes output style. In the evaluated cases, native search sometimes shifted models from concise extraction to explanatory summarization despite strict final-answer instructions. In our HotpotQA diagnostics, an affected Claude Sonnet 4 native-search run produced answers beginning with ``Based on the search results...'' in 78.1\% of predictions; among exact-match failures, 62.3\% still contained the correct answer as a substring. Table \ref{tab:verbosity_diagnostics} reports the diagnostic summary.

This matters when a semantically correct answer is unusable because it violates an exact-entity, JSON, boolean, or categorical output contract. Headline accuracy uses task-level judge/classifier scoring, while format compliance is reported separately as a diagnostic. DSG mitigates this by keeping retrieval as a structured tool response with a stable output boundary; future work with provider-side telemetry could further isolate the internal triggers.

\begin{figure*}[!t]
    \centering
    \includegraphics[width=\textwidth]{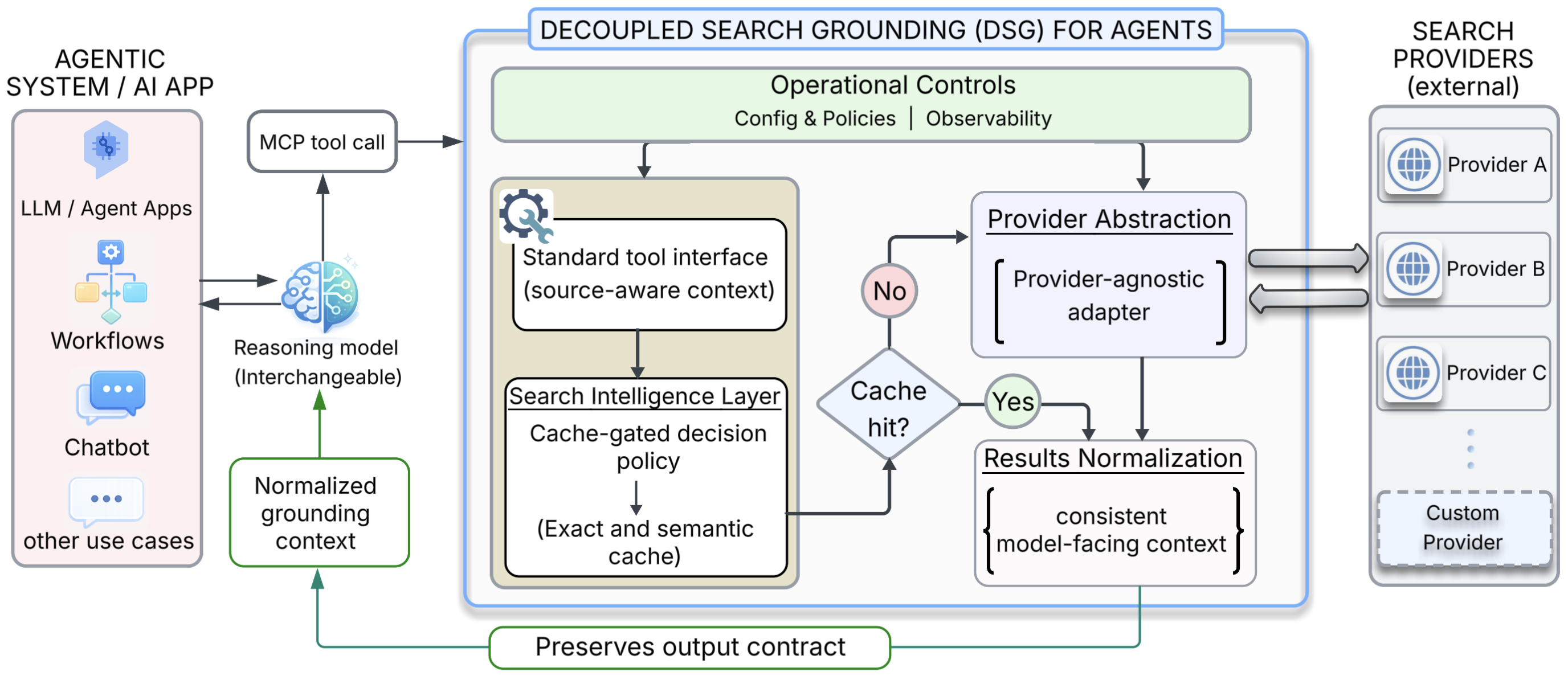}
    \caption{Decoupled Search Grounding architecture. DSG separates agentic applications and interchangeable reasoning models from external search providers through a stable tool interface, cache-gated search intelligence, provider abstraction, operational controls, and normalized model-facing grounding context.}
    \label{fig:architecture}
\end{figure*}

\section{Decoupled Search Grounding for LLM Agents}
\label{sec:architecture}

We implement Decoupled Search Grounding (DSG) as an MCP-compatible gateway between agentic applications, interchangeable reasoning models, and provider-agnostic search backends (Figure \ref{fig:architecture}).

\subsection{Source-Aware Tool Formatting}
Unlike native integrations that hide retrieval inside a provider-specific API, our architecture exposes search through explicit, configurable tool schemas. Engineers specify the query interface and provider options once at the gateway layer, so each agent inherits them without hard-coding. When the LLM emits a search tool call, DSG executes the query through its MCP-compatible interface and normalizes provider responses into a structured internal result object. The rendered tool context is consistent across providers and pairs snippets with source URLs, giving the model explicit provenance cues for which evidence to use.

These URLs also give engineers an inspectable artifact for logging, debugging, source filtering, and fallback. Returning results as a discrete tool response preserves the retrieval-generation boundary, which a provider-managed context stream collapses.

\subsection{Provider Abstraction and Fallback}
The gateway acts as a grounding control plane where provider choice, fallback order, and cost metadata are configurable policy applied across agents. The \textit{Provider Registry} normalizes heterogeneous search providers (e.g., Serper, BrightData, Firecrawl, Exa). Standard HTTP providers can be onboarded through a YAML adapter that specifies endpoint, method, headers, request templates, response-field mappings, capabilities, and cost metadata; richer providers use dedicated adapters behind the same interface. This lets researchers and engineers tune grounding for different applications, including freshness, cost, latency, or source coverage, without changing the reasoning model or application prompt. A provider timeout or error response advances the request along the configured fallback chain, so transient provider failures do not interrupt the grounding context the model receives.

\subsection{Search Intelligence Layer}
A central advantage of decoupling is that search policy can be optimized across requests, beyond the scope of a single model call. DSG introduces a search intelligence layer that applies a tiered decision policy: exact cache lookup, semantic cache reuse, and configured provider fallback for novel queries. Let $q$ denote a query, $\bar{q}$ its normalized form, $\mathcal{C}=\{(q_i,R_i)\}$ a provider-isolated semantic cache, $e(\cdot)$ an embedding function, and $p_1,\ldots,p_k$ a configured provider fallback chain. The gateway decision is:

\begin{equation}
\begin{aligned}
j^* &= \arg\max_j \cos(e(q), e(q_j)),\\
D(q)&=
\begin{cases}
R_e, & E(\bar{q})\\
R_{j^*}, & s_{j^*} \geq \tau\\
S(\mathbf{p}, q), & \text{else}
\end{cases}
\end{aligned}
\end{equation}

where $s_{j^*}=\cos(e(q),e(q_{j^*}))$, $E$ denotes an exact-cache hit, $S$ executes the configured fallback chain $\mathbf{p}=(p_1,\ldots,p_k)$, and $\tau$ is a configurable semantic-match threshold. The DSG cache architecture stores exact and semantic entries with provider-scoped keys, so cached results from one provider do not silently replace another provider's output. Cache entries also carry provider and domain specific time-to-live bounds, so recency-sensitive queries expire quickly while static factoid results persist, keeping high cache-hit rates from compromising grounding freshness. This keeps provider policy explicit: repeated queries can bypass external APIs, semantically similar queries can reuse compatible evidence, and novel queries fall through to the configured provider chain. Algorithm \ref{alg:routing} summarizes the gateway execution path.

\begin{algorithm}[t]
\small
\SetAlgoLined
\KwIn{Query $q$, threshold $\tau$, fallback chain $p_1,\ldots,p_k$}
\KwOut{Rendered tool context $T$}
 $\bar{q} \leftarrow \text{Normalize}(q)$\;
 $R \leftarrow \text{ExactCache.Get}(\bar{q})$\;
 \If{$R \neq \emptyset$}{
  \Return $\text{RenderContext}(R)$\;
 }
 $z \leftarrow \text{Embed}(q)$\;
 $(q_j, R_j, s_j) \leftarrow \text{NearestSemanticCache}(z)$\;
 \If{$s_j \geq \tau$}{
  \Return $\text{RenderContext}(R_j)$\;
 }
 \For{$p \in (p_1,\ldots,p_k)$}{
  \If{$p$ is unavailable}{continue\;}
  $R_{\mathrm{raw}} \leftarrow \text{ExecuteSearch}(p,q)$\;
  \If{$R_{\mathrm{raw}}$ has results}{
   $R \leftarrow \text{NormalizeProviderResult}(R_{\mathrm{raw}})$\;
   $\text{ExactCache.Set}(\bar{q}, R)$\;
   $\text{SemanticCache.Set}(q, z, R)$\;
   \Return $\text{RenderContext}(R)$\;
  }
 }
 \Return $\text{RenderContext}(\emptyset)$\;
\caption{Cache-Gated Provider Execution}
 \label{alg:routing}
\end{algorithm}

DSG makes grounding a shared system layer outside the model API. A provider integration can serve multiple applications and reasoning models through the same normalized result structure. Routing, fallback, caching, retrieval depth, provenance, telemetry, and cost attribution are centralized, so model or provider changes require gateway configuration or adapter updates instead of application and prompt rewrites. The stable model-facing schema also enables controlled comparisons of quality, freshness, cost, latency, and output behavior under fixed prompts. Section \ref{sec:evaluation} evaluates these controls, and Appendix \ref{sec:appendix_deployment_criteria} summarizes the resulting deployment surfaces.
\section{Empirical Evaluation}
\label{sec:evaluation}

We evaluate DSG as a systems choice: accuracy retained, cost paid, and controls exposed when search moves outside the model-provider boundary. Benchmarks span static factuality (SimpleQA; \citealp{wei2024simpleqa}), recency-sensitive knowledge (FreshQA; \citealp{vu2024freshllms}), multi-hop QA (HotpotQA; \citealp{yang2018hotpotqa}), and production-style e-commerce grounding. Public QA uses fixed samples, fixed prompts, and task-level scoring. SimpleQA and FreshQA use GPT-4.1 judging, with a 3-judge SimpleQA validation ablation showing 97.78\% mean unanimous agreement (Appendix \ref{sec:appendix_multijudge}); HotpotQA uses deterministic EM/F1. Prompt-compliance diagnostics are reported separately from headline accuracy. For QIU, we follow the multi-category marketplace intent-routing protocol of \citet{boateng2026agentic}: \textit{Retail} ($N=7{,}988$) covers general non-food retail queries, and \textit{Tail (Synthetic)} ($N=2{,}335$) covers rare long-tail queries. Detailed QIU provenance and evaluation protocol are provided in Appendix \ref{sec:appendix_datasets}. Costs are provider-side search costs per 1K queries, excluding model inference cost; latencies are benchmark latencies from evaluation outputs.

\begin{figure*}[!t]
    \centering
    \includegraphics[width=0.93\textwidth]{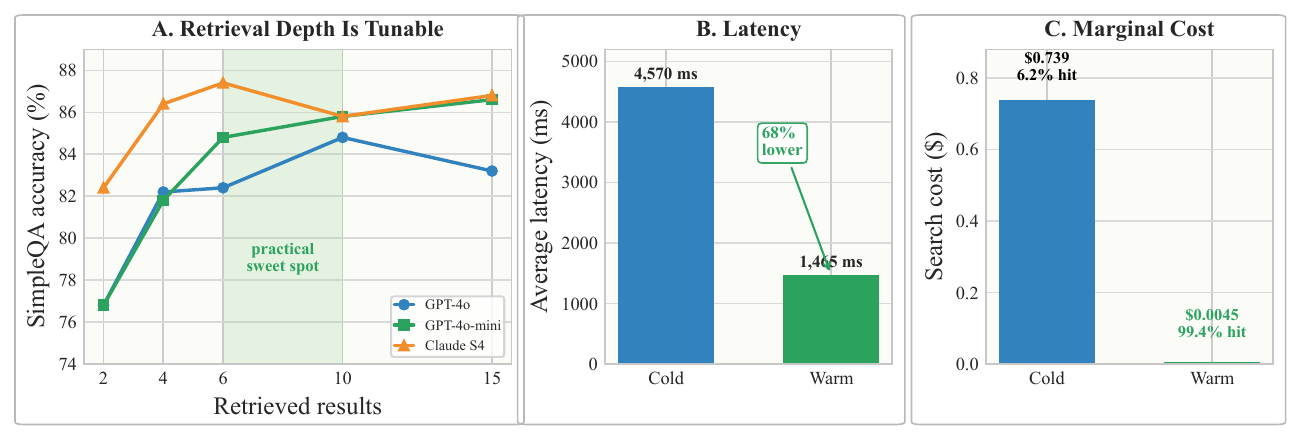}
    \caption{Operational controls enabled by DSG: retrieval-depth tuning and a controlled repeated-query warm-path replay reduce latency from 4,570ms to 1,465ms and nearly eliminate marginal search cost.}
    \label{fig:controls}
\end{figure*}

\begin{table*}[t]
\centering
\small
\setlength{\tabcolsep}{5pt}
\begin{tabular*}{\textwidth}{@{\extracolsep{\fill}}llrrrr}
\toprule
\textbf{Task} & \textbf{Scope} & \textbf{No Search} & \textbf{Native} & \textbf{DSG+BrightData} & \textbf{DSG+Serper} \\
\midrule
SimpleQA & 5-model mean & 30.8 / \$0 & \textbf{87.7} / \$20.00 & 86.1 / \$1.80 & 83.3 / \$0.67 \\
FreshQA & 5-model mean & 56.4 / \$0 & \textbf{72.6} / \$20.00 & 68.0 / \$1.86 & 67.3 / \$0.66 \\
QIU Retail & Gemini Flash & 91.10 / \$0 & 93.40 / \$7.90 & 93.81 / \$0.238 & \textbf{93.90} / \$0.110 \\
QIU Tail (Synthetic) & Gemini Flash & 83.08 / \$0 & 87.62 / \$10.37 & 86.61 / \$0.386 & \textbf{87.79} / \$0.146 \\
\bottomrule
\end{tabular*}
\caption{Main results: accuracy / search cost per 1K queries. Academic rows average five models; QIU uses Gemini Flash.}
\label{tab:main_results}
\end{table*}

\subsection{Cost-Accuracy Pareto Optimality}
\begin{figure}[t]
    \centering
    \includegraphics[width=\linewidth]{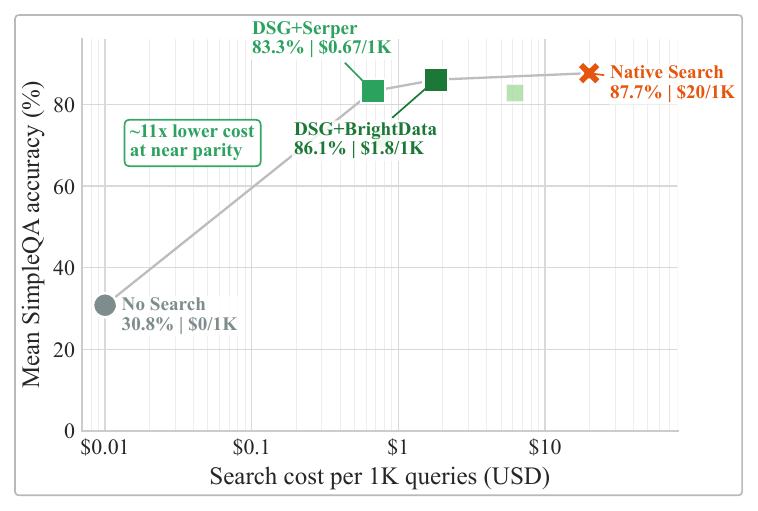}
    \caption{SimpleQA cost-accuracy frontier, averaged across five models. DSG+BrightData nearly matches native accuracy at roughly 11x lower search cost, while DSG+Serper provides the lowest-cost high-accuracy operating point.}
    \label{fig:pareto}
\end{figure}

Native integrations package retrieval quality with fixed provider pricing. Table \ref{tab:main_results} and Figure \ref{fig:pareto} show that DSG routing creates a broader operating frontier. On SimpleQA, DSG+BrightData reaches 86.1\% mean accuracy, close to native search at 87.7\%, while reducing average search-provider cost from \$20.00 to \$1.80 per 1K queries. DSG+Serper provides a lower-cost operating point at 83.3\% accuracy and \$0.67 per 1K queries. Bootstrap confidence intervals are reported in Appendix \ref{sec:appendix_simpleqa_ci}.

The trade-off is task-dependent. On FreshQA, native search leads on recency-sensitive questions, consistent with tightly integrated proprietary freshness or indexing advantages. This makes provider selection an explicit system decision across freshness, cost, inspection, portability, and caching.

\subsection{Real-World Impact: E-commerce QIU}
QIU classifies ambiguous marketplace queries using catalog and web evidence, with accuracy measured on the final resolved intent \citep{boateng2026agentic}; in our deployment, DSG serves as the shared production grounding layer for multiple applications, including the QIU model evaluated here. The native baseline uses Gemini Flash's Google Search grounding \citep{google2026gemini_search_grounding}. On Retail, DSG+Serper reaches 93.90\% versus 93.40\% for native search while reducing search cost from \$7.90 to \$0.110 per 1K queries. On Tail (Synthetic), it reaches 87.79\% versus 87.62\% at \$0.146 per 1K queries, over 98\% below native search.

\begin{figure*}[t]
    \centering
    \includegraphics[width=0.98\textwidth]{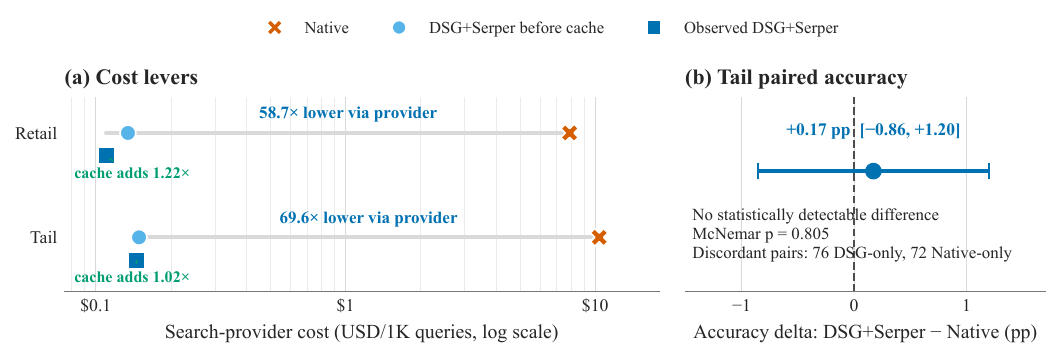}
    \caption{QIU cost decomposition and paired Tail accuracy. Provider selection explains most of the search-cost reduction, while the paired Tail interval shows no statistically detectable difference from native search. No-cache costs are estimated from observed search usage and Serper's per-search price.}
    \label{fig:qiu_decomposition}
\end{figure*}

Figure \ref{fig:qiu_decomposition} shows no statistically detectable Tail accuracy difference and that provider selection through DSG drives most of the savings while preserving the same normalized model-facing result schema across providers. Before cache reuse, DSG+Serper is 58.7$\times$ cheaper on Retail and 69.6$\times$ cheaper on Tail; caching adds 1.22$\times$ and 1.02$\times$ reductions. Detailed statistics, calculations, and provider results are reported in Appendix \ref{sec:appendix_qiu}.

\subsection{Operational Controls: Caching and Tuning}
Because DSG intercepts structured tool calls, it exposes controls unavailable in black-box search. A SimpleQA \texttt{max\_results} sweep shows quality rising sharply from 2 to 4--6 results before saturating around 6--10 (Figure \ref{fig:controls}A), turning grounding into a tunable budget knob. In a controlled replay of the same 500 queries with GPT-4o and DSG+BrightData, the warm pass reaches a 99.4\% hit rate, reduces average latency from 4,570ms to 1,465ms, and nearly eliminates marginal search-provider cost (Figure \ref{fig:controls}B--C). Holding the workload, model, and provider fixed isolates the attainable warm-cache contribution.
\section{Conclusion}
\label{sec:lessons}

We argue that real-time grounding for LLM agents should be an optimizable systems boundary that teams control directly. By separating retrieval from reasoning, DSG keeps the reasoning model interchangeable while exposing provider policy, source-aware rendering, retrieval-depth tuning, fallback, and warm-cache reuse as explicit controls. Across five frontier models and a production workload, it approaches native-search accuracy on public QA, matches native-search accuracy on QIU at far lower cost, and holds strict output contracts where native search drifts into prose in our evaluated diagnostics. The most consequential open problem we surface is Search-Induced Verbosity, where a correct answer wrapped in prose still breaks downstream parsers; future work should measure it across models with provider-side telemetry, add contract-aware validators at the boundary, and learn adaptive routing, retrieval depth, and multi-hop strategies over the same interface.

\section*{Limitations}
\label{sec:limitations}

While our DSG architecture demonstrates significant advantages in cost, latency, and prompt compliance, we acknowledge several limitations in our study and the proposed system.

\paragraph{Reliance on Third-Party APIs}
The decoupled architecture fundamentally relies on the stability, latency, and data quality of third-party search vendors (e.g., Serper, BrightData, Firecrawl). Changes to these external APIs, including pricing adjustments, rate limits, or indexing behavior, could impact the cost-accuracy Pareto frontier presented in this work. Reported provider costs reflect pricing observed during our evaluation period and are expected to change over time; our contribution is the controllable grounding interface and the operating frontiers it exposes, not any specific price point.

\paragraph{Multi-Hop Reasoning Constraints}
Our evaluation on the HotpotQA benchmark (detailed in Appendix \ref{sec:appendix_hotpotqa}) revealed that search augmentation, whether native or DSG-based, provides only modest gains for complex, multi-hop reasoning tasks compared to single-hop factoid retrieval (SimpleQA). This suggests that while our architecture effectively grounds models for direct queries, solving deep, multi-step reasoning problems likely requires more advanced retrieval-reasoning coordination, such as iterative retrieval, planning, or self-reflection \citep{trivedi2023ircot, asai2024selfrag, jeong2024adaptive}, beyond simple search decoupling. We hypothesize that DSG's tool boundary is most effective for controllable retrieval, while complex multi-hop tasks may require tighter iterative loops with planning, reflection, and repeated evidence acquisition.

\paragraph{Evaluation Methodology}
For academic benchmarks (SimpleQA, FreshQA), we utilized an LLM-as-a-judge framework (GPT-4.1). Model-based and RAG evaluation have established precedents in the literature \citep{liu2023geval, zheng2023judging, saadfalcon2024ares, es2024ragas}, while prior work reports bias toward LLM-generated text and position, verbosity, and self-enhancement effects \citep{liu2023geval, zheng2023judging}. We mitigated this by utilizing a multi-judge majority voting ablation (Appendix \ref{sec:appendix_multijudge}), but human evaluation remains the gold standard for nuanced qualitative assessment.

\section*{Ethical Considerations}
For semantic caching, the gateway is designed to cache only provider-permitted and ethically allowable context under applicable provider terms and data-use agreements, rather than indiscriminately storing retrieved content. The public results artifact releases non-proprietary public-benchmark outputs, aggregate tables, and figures while excluding proprietary QIU data, provider-restricted retrieved snippets, and internal deployment details.

\section*{Acknowledgments}
We thank Elyse Winer, Drishya Giri, Kaiwen Bian, Jamie Gu, Prabhjot Saini, Stephanie Poon, Sebastian Connelly, Tanvi Priya, Johny Rufus, and Tom Tang for their contributions, feedback, and support throughout this work. An early version of this work received the Engineering Excellence award at the 2026 DoorDash AI Hackathon.

\bibliography{references}

\appendix
\raggedbottom
\renewcommand{\topfraction}{0.95}
\renewcommand{\bottomfraction}{0.9}
\renewcommand{\textfraction}{0.05}
\renewcommand{\floatpagefraction}{0.75}
\renewcommand{\dbltopfraction}{0.95}
\renewcommand{\dblfloatpagefraction}{0.75}
\setcounter{topnumber}{4}
\setcounter{bottomnumber}{4}
\setcounter{totalnumber}{10}
\setcounter{dbltopnumber}{4}

\section{Dataset Details}
\label{sec:appendix_datasets}

To ensure a comprehensive evaluation, we utilized a mix of public academic benchmarks and proprietary industry datasets.

Native baselines were evaluated through each model vendor's native search integration, while external providers were evaluated through DSG. Public-benchmark comparisons forced search and disabled semantic-cache reuse, so their search-cost differences reflect provider selection and pricing with no contribution from cache hits.

\paragraph{SimpleQA} A benchmark designed to evaluate the factuality of LLMs on short, single-hop factoid questions. It is highly sensitive to hallucinations and requires precise entity extraction. We utilized a random subsample of $N=1,000$ questions.
\paragraph{FreshQA} A dynamic QA benchmark ($N=500$) that tests a model's ability to retrieve highly recency-sensitive knowledge (e.g., ``Who won the most recent Super Bowl?''). This dataset specifically tests the latency and freshness of the underlying search indices.
\paragraph{HotpotQA} A multi-hop reasoning dataset ($N=1,000$) where answering the question requires retrieving and synthesizing information from multiple distinct sources.
\paragraph{Proprietary QIU (Retail \& Tail (Synthetic))} Query Intent Understanding (QIU) datasets sourced from a large-scale e-commerce platform. The \textit{Retail} dataset ($N=7,988$) covers general non-food retail queries, while the \textit{Tail (Synthetic)} dataset ($N=2,335$) covers rare long-tail queries. These datasets represent the complex, noisy inputs typical of real-world marketplace search. Both are fixed evaluation snapshots dated November 14, 2025, and were not used to train or tune DSG, its adapters, or routing policy. Labels were withheld from all inference inputs and joined only for post-inference scoring; every configuration used the same query-label sets.

\section{Visual Summary of Full Results}
\label{sec:appendix_visual_summary}

\begin{figure*}[t]
    \centering
    \includegraphics[width=\textwidth]{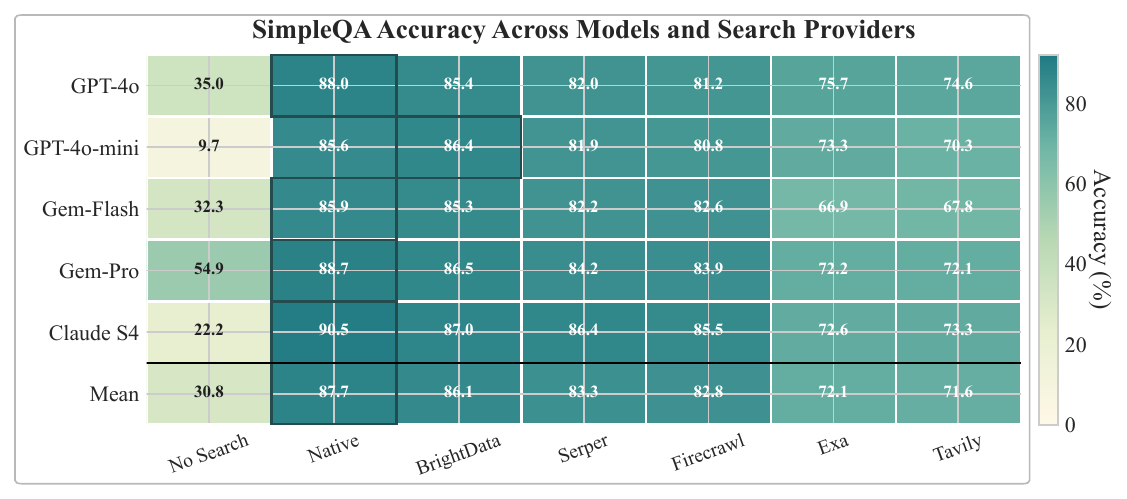}
    \caption{SimpleQA accuracy heatmap across all evaluated models and search providers. Outlined cells mark the best configuration for each row; the mean row summarizes the overall provider pattern.}
    \label{fig:appendix_simpleqa_heatmap}
\end{figure*}

\begin{figure*}[t]
    \centering
    \includegraphics[width=\textwidth]{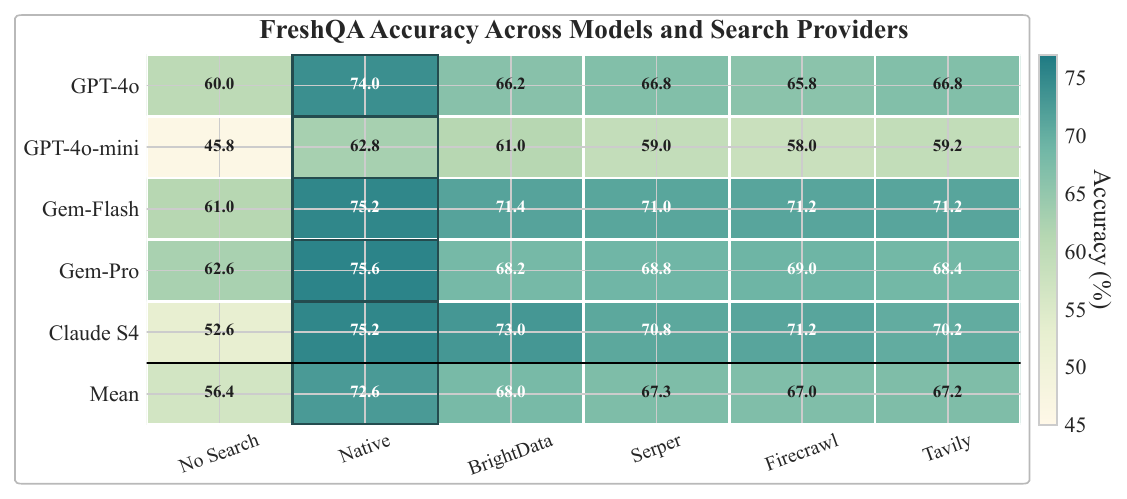}
    \caption{FreshQA accuracy heatmap. Outlined cells mark the best configuration for each row. Native search leads across all evaluated models on this recency-sensitive benchmark; Exa was not run for FreshQA.}
    \label{fig:appendix_freshqa_heatmap}
\end{figure*}

\begin{figure*}[t]
    \centering
    \includegraphics[width=\textwidth]{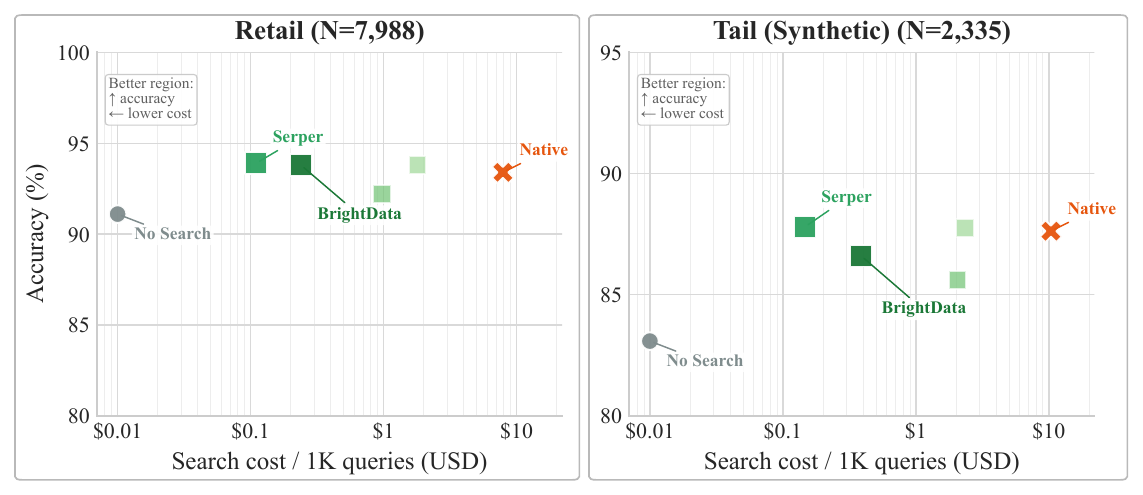}
    \caption{QIU cost-accuracy trade-offs for Retail and Tail (Synthetic). Exact provider results are reported in Table \ref{tab:qiu_provider_full}.}
    \label{fig:appendix_qiu_tradeoff}
\end{figure*}

Figures \ref{fig:appendix_simpleqa_heatmap}--\ref{fig:appendix_qiu_tradeoff} provide compact visual summaries of the comprehensive tables below. The plots are intended for fast pattern inspection; the exact values and caveats remain in the corresponding tables.

\section{Comprehensive Results: SimpleQA}
\label{sec:appendix_comprehensive}

Table \ref{tab:simpleqa_full} presents the comprehensive results on the SimpleQA benchmark ($N=1,000$) across all evaluated models and search providers. This expands upon the summary presented in Section \ref{sec:evaluation}.

\begin{table*}[t]
\centering
\small
\begin{tabular}{llrrrr}
\toprule
\textbf{Model} & \textbf{Search Config} & \textbf{Acc (\%)} & \textbf{Cost (\$)} & \textbf{Median Latency (ms)} & \textbf{P95 (ms)} \\
\midrule
\multirow{7}{*}{GPT-4o} & No Search & 35.0 & 0.00 & 530 & 1012 \\
& Native & \textbf{88.0} & 10.00 & 2717 & 4543 \\
& DSG+BrightData & 85.4 & 1.54 & 3913 & 9595 \\
& DSG+Serper & 82.0 & 0.53 & 2119 & 3696 \\
& DSG+Firecrawl & 81.2 & 1.81 & 2563 & 5566 \\
& DSG+Exa & 75.7 & 0.55 & 1804 & 8920 \\
& DSG+Tavily & 74.6 & 0.54 & 2646 & 4933 \\
\midrule
\multirow{7}{*}{GPT-4o-mini} & No Search & 9.7 & 0.00 & 526 & 978 \\
& Native & 85.6 & 10.00 & 1458 & 2255 \\
& DSG+BrightData & \textbf{86.4} & 2.00 & 4087 & 12212 \\
& DSG+Serper & 81.9 & 0.76 & 2434 & 6125 \\
& DSG+Firecrawl & 80.8 & 22.62 & 3201 & 8397 \\
& DSG+Exa & 73.3 & 0.90 & 2729 & 7286 \\
& DSG+Tavily & 70.3 & 0.93 & 4258 & 9823 \\
\midrule
\multirow{7}{*}{Gemini 2.5 Flash} & No Search & 32.3 & 0.00 & 1610 & 3699 \\
& Native & \textbf{85.9} & 35.00 & 3110 & 8049 \\
& DSG+BrightData & 85.3 & 1.66 & 5836 & 18116 \\
& DSG+Serper & 82.2 & 0.59 & 3822 & 9550 \\
& DSG+Firecrawl & 82.6 & 1.94 & 4307 & 13830 \\
& DSG+Exa & 66.9 & 0.69 & 3065 & 11621 \\
& DSG+Tavily & 67.8 & 0.68 & 3143 & 11389 \\
\midrule
\multirow{7}{*}{Gemini 2.5 Pro} & No Search & 54.9 & 0.00 & 5989 & 12207 \\
& Native & \textbf{88.7} & 35.00 & 5063 & 13609 \\
& DSG+BrightData & 86.5 & 1.80 & 9107 & 27629 \\
& DSG+Serper & 84.2 & 0.67 & 6539 & 19139 \\
& DSG+Firecrawl & 83.9 & 2.17 & 7976 & 23464 \\
& DSG+Exa & 72.2 & 0.76 & 7853 & 23135 \\
& DSG+Tavily & 72.1 & 0.78 & 8165 & 24551 \\
\midrule
\multirow{7}{*}{Claude Sonnet 4} & No Search & 22.2 & 0.00 & 1294 & 2482 \\
& Native & \textbf{90.5} & 9.99 & 4712 & 15472 \\
& DSG+BrightData & 87.0 & 2.01 & 6380 & 17853 \\
& DSG+Serper & 86.4 & 0.79 & 4281 & 10894 \\
& DSG+Firecrawl & 85.5 & 2.58 & 4702 & 12750 \\
& DSG+Exa & 72.6 & 0.94 & 4668 & 11248 \\
& DSG+Tavily & 73.3 & 0.96 & 6837 & 13817 \\
\bottomrule
\end{tabular}
\caption{Comprehensive SimpleQA results detailing accuracy, cost, and latency across all tested configurations.}
\label{tab:simpleqa_full}
\end{table*}

\section{Comprehensive Results: FreshQA}
\label{sec:appendix_freshqa}

Table \ref{tab:freshqa_full} presents the comprehensive results on the FreshQA benchmark ($N=500$), which tests recency-sensitive knowledge. Native search is the accuracy leader across all five models; Exa was not run for FreshQA and is therefore omitted.

\begin{table*}[t]
\centering
\small
\begin{tabular}{llrrrr}
\toprule
\textbf{Model} & \textbf{Search Config} & \textbf{Acc (\%)} & \textbf{Cost (\$)} & \textbf{Median Latency (ms)} & \textbf{P95 (ms)} \\
\midrule
\multirow{6}{*}{GPT-4o} & No Search & 60.0 & 0.00 & 463 & 1043 \\
& Native & \textbf{74.0} & 9.98 & 2558 & 4179 \\
& DSG+BrightData & 66.2 & 1.55 & 3955 & 10157 \\
& DSG+Serper & 66.8 & 0.52 & 1976 & 3172 \\
& DSG+Firecrawl & 65.8 & 1.78 & 1501 & 3122 \\
& DSG+Tavily & 66.8 & 0.53 & 2225 & 3600 \\
\midrule
\multirow{6}{*}{GPT-4o-mini} & No Search & 45.8 & 0.00 & 468 & 883 \\
& Native & \textbf{62.8} & 10.00 & 1717 & 2913 \\
& DSG+BrightData & 61.0 & 2.15 & 4705 & 13601 \\
& DSG+Serper & 59.0 & 0.77 & 2516 & 6128 \\
& DSG+Firecrawl & 58.0 & 2.68 & 1866 & 6161 \\
& DSG+Tavily & 59.2 & 0.75 & 2530 & 6795 \\
\midrule
\multirow{6}{*}{Gemini 2.5 Flash} & No Search & 61.0 & 0.00 & 1348 & 3323 \\
& Native & \textbf{75.2} & 35.00 & 2796 & 8254 \\
& DSG+BrightData & 71.4 & 1.64 & 4837 & 16012 \\
& DSG+Serper & 71.0 & 0.60 & 2740 & 7950 \\
& DSG+Firecrawl & 71.2 & 1.97 & 2166 & 8324 \\
& DSG+Tavily & 71.2 & 0.58 & 2831 & 8076 \\
\midrule
\multirow{6}{*}{Gemini 2.5 Pro} & No Search & 62.6 & 0.00 & 5555 & 22210 \\
& Native & \textbf{75.6} & 35.00 & 5531 & 18083 \\
& DSG+BrightData & 68.2 & 1.82 & 8310 & 31556 \\
& DSG+Serper & 68.8 & 0.62 & 6051 & 28749 \\
& DSG+Firecrawl & 69.0 & 2.06 & 5919 & 21108 \\
& DSG+Tavily & 68.4 & 0.61 & 6598 & 29817 \\
\midrule
\multirow{6}{*}{Claude Sonnet 4} & No Search & 52.6 & 0.00 & 1182 & 2867 \\
& Native & \textbf{75.2} & 10.00 & 5725 & 17726 \\
& DSG+BrightData & 73.0 & 2.16 & 6336 & 16846 \\
& DSG+Serper & 70.8 & 0.81 & 4276 & 12447 \\
& DSG+Firecrawl & 71.2 & 2.53 & 3723 & 11037 \\
& DSG+Tavily & 70.2 & 0.81 & 4116 & 11567 \\
\bottomrule
\end{tabular}
\caption{Comprehensive FreshQA results. Evaluated using GPT-4.1 judge in RELAXED mode.}
\label{tab:freshqa_full}
\end{table*}

\section{Comprehensive Results: HotpotQA}
\label{sec:appendix_hotpotqa}

Table \ref{tab:hotpotqa_full} presents the comprehensive results on the HotpotQA benchmark ($N=1,000$). Unlike SimpleQA, which tests single-hop factoid retrieval, HotpotQA requires multi-hop reasoning. Search augmentation provides modest and model-dependent gains, but DSG+BrightData is the strongest overall DSG configuration and is competitive with or better than native search for most models. All rows are reported with task-level EM/F1 scoring.

\begin{table*}[t]
\centering
\small
\begin{tabular}{llrrrrr}
\toprule
\textbf{Model} & \textbf{Search Config} & \textbf{EM (\%)} & \textbf{F1} & \textbf{Cost (\$)} & \textbf{Median Latency (ms)} & \textbf{P95 (ms)} \\
\midrule
\multirow{7}{*}{GPT-4o} & No Search & 38.1 & 0.521 & 0.00 & 502 & 1127 \\
& Native & 37.6 & 0.541 & 10.00 & 2999 & 5860 \\
& DSG+BrightData & 37.4 & 0.545 & 1.97 & 4881 & 12339 \\
& DSG+Serper & 35.6 & 0.519 & 0.68 & 2364 & 4699 \\
& DSG+Firecrawl & 37.4 & 0.530 & 2.32 & 2937 & 6462 \\
& DSG+Exa & 34.7 & 0.515 & 0.67 & 2088 & 4139 \\
& DSG+Tavily & 36.3 & 0.529 & 0.65 & 2667 & 5110 \\
\midrule
\multirow{7}{*}{GPT-4o-mini} & No Search & 29.3 & 0.404 & 0.00 & 487 & 835 \\
& Native & 39.6 & 0.523 & 10.00 & 1477 & 2297 \\
& DSG+BrightData & \textbf{41.0} & \textbf{0.566} & 2.51 & 5811 & 14773 \\
& DSG+Serper & 39.8 & 0.548 & 0.89 & 3331 & 5893 \\
& DSG+Firecrawl & 39.0 & 0.539 & 3.04 & 4499 & 8049 \\
& DSG+Exa & 37.6 & 0.527 & 0.86 & 2710 & 6641 \\
& DSG+Tavily & 36.8 & 0.521 & 0.85 & 4033 & 8368 \\
\midrule
\multirow{7}{*}{Gemini 2.5 Flash} & No Search & 41.1 & 0.537 & 0.00 & 1873 & 8265 \\
& Native & 42.0 & 0.578 & 35.00 & 3110 & 18943 \\
& DSG+BrightData & \textbf{44.7} & \textbf{0.591} & 2.36 & 11482 & 25481 \\
& DSG+Serper & 44.2 & 0.582 & 0.81 & 5484 & 13570 \\
& DSG+Firecrawl & 44.0 & 0.572 & 2.74 & 5587 & 16691 \\
& DSG+Exa & 43.1 & 0.576 & 0.78 & 5053 & 13261 \\
& DSG+Tavily & 43.9 & 0.577 & 0.79 & 5475 & 13895 \\
\midrule
\multirow{7}{*}{Gemini 2.5 Pro} & No Search & 44.7 & 0.596 & 0.00 & 5923 & 19550 \\
& Native & 45.1 & 0.610 & 35.00 & 6798 & 23193 \\
& DSG+BrightData & \textbf{47.7} & \textbf{0.627} & 2.43 & 12321 & 37805 \\
& DSG+Serper & 46.4 & 0.610 & 50.86 & 9263 & 29404 \\
& DSG+Firecrawl & 46.4 & 0.608 & 2.85 & 10518 & 30059 \\
& DSG+Exa & 46.1 & 0.606 & 0.78 & 8872 & 41160 \\
& DSG+Tavily & 45.6 & 0.602 & 0.79 & 9612 & 37290 \\
\midrule
\multirow{7}{*}{Claude Sonnet 4} & No Search & 33.1 & 0.442 & 0.00 & 1264 & 3189 \\
& Native & 32.6 & 0.444 & 10.00 & 6761 & 17811 \\
& DSG+BrightData & \textbf{43.3} & \textbf{0.581} & 2.94 & 8777 & 19843 \\
& DSG+Serper & 43.3 & 0.571 & 1.04 & 5439 & 10621 \\
& DSG+Firecrawl & 41.1 & 0.551 & 3.45 & 7711 & 16034 \\
& DSG+Exa & 39.7 & 0.525 & 36.00 & 4607 & 10549 \\
& DSG+Tavily & 39.2 & 0.521 & 36.00 & 6657 & 12848 \\
\bottomrule
\end{tabular}
\caption{Comprehensive HotpotQA results. Exact Match (EM) and F1 scores are deterministic. Results are more mixed than SimpleQA, reflecting the additional difficulty of multi-hop retrieval and exact-answer formatting.}
\label{tab:hotpotqa_full}
\end{table*}

\section{QIU Provider Tables}
\label{sec:appendix_qiu}

The QIU datasets are proprietary e-commerce intent-understanding evaluations. Table \ref{tab:qiu_provider_full} reports the final provider comparison used in the main analysis.

\begin{table*}[!t]
\centering
\scriptsize
\textbf{Tail (Synthetic) ($N=2{,}335$)}\\[-0.5mm]
\resizebox{\textwidth}{!}{
\begin{tabular}{lrrrrr}
\toprule
\textbf{Config} & \textbf{Acc (\%)} & \textbf{$\Delta$ vs. No Search} & \textbf{Search Usage (\%)} & \textbf{Cache (\%)} & \textbf{Cost (\$/1K)} \\
\midrule
No Search & 83.08 & -- & 0.0 & 0.0 & 0.000 \\
Google (Native) & 87.62 & +4.54pp & 29.6 & 0.0 & 10.373 \\
DSG+Tavily & 87.75 & +4.67pp & 29.8 & 1.15 & 2.354 \\
DSG+Serper & \textbf{87.79} & +4.71pp & 29.8 & 1.73 & \textbf{0.146} \\
DSG+BrightData & 86.61 & +3.53pp & 26.9 & 3.19 & 0.386 \\
DSG+Firecrawl & 85.61 & +2.53pp & 26.9 & 4.31 & 2.056 \\
\bottomrule
\end{tabular}
}
\vspace{1.5mm}

\textbf{Retail ($N=7{,}988$)}\\[-0.5mm]
\resizebox{\textwidth}{!}{
\begin{tabular}{lrrrrr}
\toprule
\textbf{Config} & \textbf{Acc (\%)} & \textbf{$\Delta$ vs. No Search} & \textbf{Search Usage (\%)} & \textbf{Cache (\%)} & \textbf{Cost (\$/1K)} \\
\midrule
No Search & 91.10 & -- & 0.0 & 0.0 & 0.000 \\
Google (Native) & 93.40 & +2.30pp & 22.5 & 0.0 & 7.900 \\
DSG+Tavily & 93.80 & +2.70pp & 26.9 & 14.4 & 1.800 \\
DSG+Serper & \textbf{93.90} & +2.80pp & 26.9 & 14.9 & \textbf{0.110} \\
DSG+BrightData & 93.81 & +2.71pp & 23.2 & 25.53 & 0.238 \\
DSG+Firecrawl & 92.20 & +1.10pp & 23.2 & 26.77 & 0.969 \\
\bottomrule
\end{tabular}
}
\caption{QIU provider comparisons for Tail (Synthetic) and Retail.}
\label{tab:qiu_provider_full}
\end{table*}

\begin{center}
\centering
\small
\resizebox{\linewidth}{!}{
\begin{tabular}{lrrrrr}
\toprule
\textbf{Split} & \textbf{Native} & \textbf{No-cache DSG} & \textbf{Observed DSG} & \textbf{Provider} & \textbf{Cache} \\
 & \textbf{(\$/1K)} & \textbf{(\$/1K)} & \textbf{(\$/1K)} & \textbf{reduction} & \textbf{reduction} \\
\midrule
Retail & 7.900 & 0.1345 & 0.110 & 58.7$\times$ & 1.22$\times$ \\
Tail (Synthetic) & 10.373 & 0.149 & 0.146 & 69.6$\times$ & 1.02$\times$ \\
\bottomrule
\end{tabular}
}
\captionof{table}{QIU search-cost decomposition. No-cache DSG+Serper is estimated from observed search usage and the provider's \$0.0005 per-search price. Provider reduction compares native with no-cache DSG; cache reduction compares no-cache with observed DSG.}
\label{tab:qiu_cost_decomposition}
\end{center}

\paragraph{Paired Tail analysis}
On the 2,335 paired Tail examples, DSG+Serper reaches 87.79\% accuracy versus 87.62\% for native search, a difference of $+0.17$ percentage points. The 95\% paired-bootstrap confidence interval is $[-0.86,+1.20]$ percentage points, and the exact McNemar test gives $p=0.805$. Discordant outcomes are nearly symmetric: 76 examples are correct only under DSG+Serper and 72 only under native search. The comparison shows no statistically detectable difference and provides no evidence of superiority.

\section{Ablations and Diagnostics}
\label{sec:appendix_ablations}

\subsection{SimpleQA Bootstrap Uncertainty}
\label{sec:appendix_simpleqa_ci}
Table \ref{tab:simpleqa_ci} reports bootstrap confidence intervals for the SimpleQA frontier summarized in Figure \ref{fig:pareto}. We resample queries with replacement within each model/configuration, compute model-level accuracy, and then average across the five evaluated models for each bootstrap replicate.

\begin{center}
\centering
\small
\begin{tabular}{lrr}
\toprule
\textbf{Configuration} & \textbf{Mean Acc. (\%)} & \textbf{95\% CI} \\
\midrule
No Search & 30.82 & [29.62, 32.06] \\
Native & 87.74 & [86.82, 88.64] \\
DSG+BrightData & 86.12 & [85.18, 87.06] \\
DSG+Serper & 83.34 & [82.32, 84.34] \\
\bottomrule
\end{tabular}
\captionof{table}{Bootstrap uncertainty for SimpleQA model-mean accuracy. Intervals use 10,000 bootstrap replicates over query-level correctness from the primary single-judge results.}
\label{tab:simpleqa_ci}
\end{center}

\subsection{Retrieval Tuning and Cache Efficiency}
Table \ref{tab:ablation_summary} reports the two operational ablations summarized in Figure \ref{fig:controls}: retrieval-depth tuning and repeated-query caching.

\begin{center}
\centering
\small
\begin{minipage}[t]{\linewidth}
\centering
\begin{tabular}{lrrr}
\toprule
\textbf{\texttt{max\_results}} & \textbf{GPT-4o} & \textbf{GPT-4o-mini} & \textbf{Claude S4} \\
\midrule
2 & 76.8\% & 76.8\% & 82.4\% \\
4 & 82.2\% & 81.8\% & 86.4\% \\
6 & 82.4\% & 84.8\% & \textbf{87.4\%} \\
10 & \textbf{84.8\%} & 85.8\% & 85.8\% \\
15 & 83.2\% & \textbf{86.6\%} & 86.8\% \\
\bottomrule
\end{tabular}
\end{minipage}
\vspace{1mm}

\begin{minipage}[t]{\linewidth}
\centering
\resizebox{\linewidth}{!}{
\begin{tabular}{lrrrr}
\toprule
\textbf{Pass} & \textbf{Hits} & \textbf{Hit (\%)} & \textbf{Cost (\$)} & \textbf{Latency (ms)} \\
\midrule
Cold & 31 & 6.2 & 0.7385 & 4570 \\
Warm & 497 & 99.4 & 0.0045 & 1465 \\
\bottomrule
\end{tabular}
}
\end{minipage}
\captionof{table}{Operational ablations. Top: effect of \texttt{max\_results} on SimpleQA accuracy with DSG+BrightData. Bottom: controlled repeated-query warm-path replay on SimpleQA ($N=500$), where the warm pass nearly eliminates marginal search cost and reduces average benchmark latency by 68\%. This holds workload, model, and provider fixed.}
\label{tab:ablation_summary}
\label{tab:max_results}
\label{tab:cache_experiment}
\end{center}

\subsection{Multi-Judge Validation}
\label{sec:appendix_multijudge}
To ensure the reliability of our single-judge LLM evaluation (GPT-4.1) on SimpleQA, we conducted a multi-judge validation ablation using a 3-judge majority verdict. The mean 3-judge unanimous agreement was 97.78\% (range 95.9\%--98.6\%). The 3-judge majority provided a slight uplift (+1.15pp) over the single judge, confirming that our primary evaluation metric is highly stable and slightly conservative.

\subsection{Prompt-Compliance Diagnostics}
\label{sec:appendix_verbosity_diagnostics}
Table \ref{tab:verbosity_diagnostics} summarizes the Claude Sonnet 4 HotpotQA diagnostic shown in Figure \ref{fig:verbosity}. The diagnostic measures the specific format-drift pattern discussed in Section \ref{sec:challenge}; it is separate from the final HotpotQA benchmark table in Appendix \ref{sec:appendix_hotpotqa}, which uses task-level EM/F1 scoring.

\begin{center}
\centering
\small
\resizebox{\linewidth}{!}{
\begin{tabular}{lrrr}
\toprule
\textbf{Mode} & \textbf{Med. Len.} & \textbf{Based... (\%)} & \textbf{Extractable (\%)} \\
\midrule
No Search & 13 & 0.0 & -- \\
DSG & 13 & 0.0 & -- \\
Native & 353 & 78.1 & 62.3 \\
\bottomrule
\end{tabular}
}
\captionof{table}{Prompt-compliance diagnostic for Claude Sonnet 4 on HotpotQA. Length is measured in output characters. ``Starts Based...'' is the fraction of predictions beginning with explanatory search-result framing; ``Extractable Failures'' is the fraction of exact-match failures that still contained the target answer as a substring.}
\label{tab:verbosity_diagnostics}
\end{center}

We treat Search-Induced Verbosity as a deployment-risk diagnostic rather than a headline accuracy metric. Headline accuracy uses task-level scoring, while this diagnostic isolates whether the final response obeys the requested output contract. The diagnostic instead isolates a production-interface risk: in workflows that require exact entities, JSON, or categorical flags, an answer can be present but wrapped in prose that breaks downstream parsers. DSG mitigates this risk by keeping retrieval as a structured tool response with a stable output boundary.

\paragraph{Additional Gemini example.}
The same pattern appears in Gemini 2.5 Flash with native grounding (Table \ref{tab:gemini_verbosity_example}). The model has the correct answer, but activating native search shifts the final response into explanatory prose that fails the exact-output contract.

\begin{center}
\centering
\small
\resizebox{\linewidth}{!}{
\begin{tabular}{p{0.26\linewidth}p{0.66\linewidth}}
\toprule
\textbf{Field} & \textbf{Example} \\
\midrule
Instruction & Provide ONLY the final answer entity. \\
Question & Which is larger, Asante Traditional Buildings or Gulangyu? \\
Target & Gulangyu \\
No Search & ``Gulangyu'' (correct) \\
Native Search & ``Gulangyu is larger. Gulangyu is an island with an area of approximately 1.88 sq...'' (fails exact match) \\
\bottomrule
\end{tabular}
}
\captionof{table}{Additional Search-Induced Verbosity example for Gemini 2.5 Flash.}
\label{tab:gemini_verbosity_example}
\end{center}

\section{System Prompts and Reproducibility}
\label{sec:appendix_prompts}

A central observation of this paper is that, in the cases we evaluated, native search integrations can weaken adherence to strict output-format instructions. To ensure reproducibility and transparency, we provide the exact system prompt utilized across all models during the factoid retrieval (SimpleQA) and multi-hop (HotpotQA) evaluations.

\begin{tcolorbox}[breakable,colback=gray!5!white,colframe=gray!75!black,title=\textbf{Standardized System Prompt}]
\small
\texttt{You are a highly precise answering agent. Your task is to answer the user's question as concisely as possible. \\
\\
CRITICAL INSTRUCTIONS:\\
1. Provide ONLY the final answer entity or short phrase.\\
2. DO NOT include any conversational filler (e.g., "The answer is...").\\
3. DO NOT explain your reasoning.\\
4. If you must use a search tool to find the answer, do so, but your final output to the user must still be ONLY the exact entity.}
\end{tcolorbox}

Despite these explicit, capitalized instructions, in our runs some models using native search (e.g., Claude Sonnet 4 with \texttt{web\_search}) often returned paragraphs of text beginning with \textit{"Based on the search results..."}. In contrast, DSG processes search results as a distinct tool response rendered from structured provider-normalized results, allowing the LLM to maintain its persona and adhere strictly to the formatting constraints.

\subsection{Source-Aware Context Rendering}
\label{sec:appendix_source_context}
DSG normalizes heterogeneous search responses into a common evidence record with fields such as \texttt{title}, \texttt{url}, and \texttt{content}. The model-facing rendering preserves this structure as numbered snippets with explicit source URLs, rather than merging retrieved text into an opaque context block. The public benchmark outputs report predictions and aggregate metrics, but omit raw retrieved snippets.

\begin{table}[t]
\centering
\small
\begin{tabular}{p{0.16\linewidth}p{0.72\linewidth}}
\toprule
\textbf{Field} & \textbf{Illustrative DSG tool context} \\
\midrule
Query & Which is larger, Asante Traditional Buildings or Gulangyu? \\
Result 1 & Title: Kulangsu, a Historic International Settlement; Source: \url{https://whc.unesco.org/en/list/1541/}; Content: UNESCO World Heritage page for Kulangsu / Gulangyu. \\
Result 2 & Title: Asante Traditional Buildings; Source: \url{https://whc.unesco.org/en/list/35/}; Content: UNESCO World Heritage page for Asante Traditional Buildings in Ghana. \\
\bottomrule
\end{tabular}
\caption{Illustrative source-aware DSG tool context.}
\label{tab:source_context_example}
\end{table}

Qualitative inspection suggests that preserving titles and URLs helps the model distinguish primary-source evidence from generic web snippets and prioritize evidence by source provenance. This source-aware rendering gives the reasoning model consistent cues about where each piece of evidence came from.

\section{Deployment Criteria Comparison}
\label{sec:appendix_deployment_criteria}

\begin{center}
\centering
\small
\resizebox{\linewidth}{!}{
\begin{tabular}{p{0.22\linewidth}p{0.34\linewidth}p{0.38\linewidth}}
\toprule
\textbf{Dimension} & \textbf{Native search} & \textbf{DSG} \\
\midrule
Accuracy & Strong, especially on freshness-sensitive queries & Near-parity on public QA; matches native on the evaluated QIU workload \\
Search cost & Fixed provider pricing & Tunable provider routing; lower search-provider cost in our evaluations \\
Latency & Provider-managed and opaque & Task/provider-dependent; cache-optimizable warm path \\
Repeated-query reuse & Not exposed & Exact + semantic cache; 99.4\% controlled warm-replay hit rate (Table \ref{tab:cache_experiment}) \\
Portability / inspectability & Coupled to model provider & Provider-agnostic registry with source-aware title/URL/snippet rendering \\
Output-contract reliability & Can induce verbose outputs & Lower format drift in diagnostics (Table \ref{tab:verbosity_diagnostics}) \\
\bottomrule
\end{tabular}
}
\captionof{table}{Expanded deployment-criteria comparison between native search and DSG. Accuracy is not assumed to favor one interface universally: native search is strongest on freshness-sensitive QA, while DSG matches native on the evaluated QIU workload and exposes deployment controls unavailable in native integrations.}
\label{tab:deployment_criteria}
\end{center}

\end{document}